\documentclass[conference]{IEEEtran}
\IEEEoverridecommandlockouts
\usepackage{cite}  
\usepackage{algorithmic}
\usepackage{graphicx}
\usepackage{textcomp}
\usepackage{xcolor}
\def\BibTeX{{\rm B\kern-.05em{\sc i\kern-.025em b}\kern-.08em
    T\kern-.1667em\lower.7ex\hbox{E}\kern-.125emX}}

  \usepackage{booktabs} 

\usepackage{amsmath,amsfonts,amsthm, mathabx, amstext}
\usepackage{dsfont, bbm,graphicx, dsfont, tabularx, colortbl,url}
\usepackage[shortlabels]{enumitem}
\usepackage{placeins} 
 \usepackage{stfloats} 
\usepackage{balance} 
\usepackage{float} 

\definecolor{LightGray}{gray}{0.9} 

\usepackage{array,multirow, colortbl}
\newcommand{\STAB}[1]{\begin{tabular}{@{}c@{}}#1\end{tabular}} 

\newcommand\numberthis{\addtocounter{equation}{1}\tag{\theequation}}

\DeclareMathOperator*{\argmax}{argmax}

\newtheorem{thm}{Theorem}

\newcounter{factnum}
\newcounter{claimnum}
 \newcounter{defnum}
\usepackage{caption}

\newcommand{\varletter}{v}
\newcommand{\querychar}{*} 
\newcommand{\querydata}{\mathcal{D}_{\querychar{}}}

\newcommand{\papertitle}{Recognizing Variables from their Data via Deep Embeddings of Distributions}

\begin{document}

\title{\papertitle}

\author{\IEEEauthorblockN{ Jonas Mueller }
\IEEEauthorblockA{\textit{ Amazon Web Services }  \\[-0.09em]
jonasmue@amazon.com  }
\and
\IEEEauthorblockN{ Alex Smola }
\IEEEauthorblockA{\textit{Amazon Web Services } \\[-0.09em]
smola@amazon.com } 
} 

\maketitle

\begin{abstract}

A key obstacle in automated analytics and meta-learning is the inability to recognize when different datasets contain measurements of the same variable.  
Because provided attribute labels are often uninformative in practice, this task may be more robustly addressed by leveraging the data values themselves rather than just relying on their arbitrarily selected variable names.
Here, we present a computationally efficient method to identify high-confidence variable matches between a given set of data values and a large repository of previously encountered datasets.  
Our approach enjoys numerous advantages over distributional similarity based techniques because we leverage learned vector embeddings of datasets which adaptively account for natural forms of data variation encountered in practice.
Based on the neural architecture of deep sets, our embeddings can be computed for both numeric and string data.
In dataset search and schema matching tasks, our methods outperform standard statistical techniques and we find that the learned embeddings generalize well to new data sources.  
  
\end{abstract}

\section{Introduction}
\label{sec:intro}

Emerging ideas in automated analytics \cite{autostatistician} and meta-learning across many datasets \cite{metalearn} offer great promise for improving both performance and tedium in the data science pipeline.  However, a major obstacle remains: such methods generally have no knowledge about what type of real-world entity (i.e.\ variable) generated the data they operate upon.  In contrast, human analysts presented with new data often utilize this knowledge to recall previously-encountered datasets that contain the same sort of variables.  Reviewing past experience with how different algorithms fared on these same variables enables a person to quickly leverage methods that work well for this type of data (e.g.\ removal of User-ID from  predictive features, missingness-modeling for user-ratings data, etc.). 

Endowing a machine with this capability can be achieved via two ingredients: (1) an organized repository full of different datasets annotated with informative metadata regarding the performance of various computational techniques, and (2) the ability to recognize which repository datasets (and hence the algorithms associated with them) are relevant when presented with new data.  
The former component is being addressed by the emergence of rapidly-growing public  repositories such as OpenML \cite{openml} and \url{kaggle.com}, in which datasets from diverse domains are curated along with the performance of thousands of data science pipelines (including preprocessing \& learning methods).  Thus, this work aims to address the latter issue of identifying repository datasets which contain some of the same variables present in newly acquired data.  

We focus on structured data tables (where rows = statistically independent observations, columns = variables), which are the most prevalent form of data among current organizations.  
Given a previously-unseen data table, we aim to automatically infer what type of variable was measured to generate the observations in each column.  To accomplish this task, we compare the new data against a stored repository of data tables and report which variables in the repository are  likely the same as variables that generated the new data.

\noindent \emph{Note on terminology: } Throughout, we use the term \emph{variable} to refer to a real-world entity whose measurements form the values contained within a single column of a data table (also commonly called an \emph{attribute} or \emph{feature}).  Although a single table of measurements (as might be found in a relational database) is often called a \emph{dataset}, we carefully refer to this as a \emph{data table}, reserving the term \emph{dataset} for the univariate data values located within a single column of the data table (i.e.\ each \emph{dataset} in this work is a single data column from some table, whose values stem from a single variable).

\begin{figure}[tb] \centering
 \vspace*{-0pt}
\includegraphics[width=0.42\textwidth]{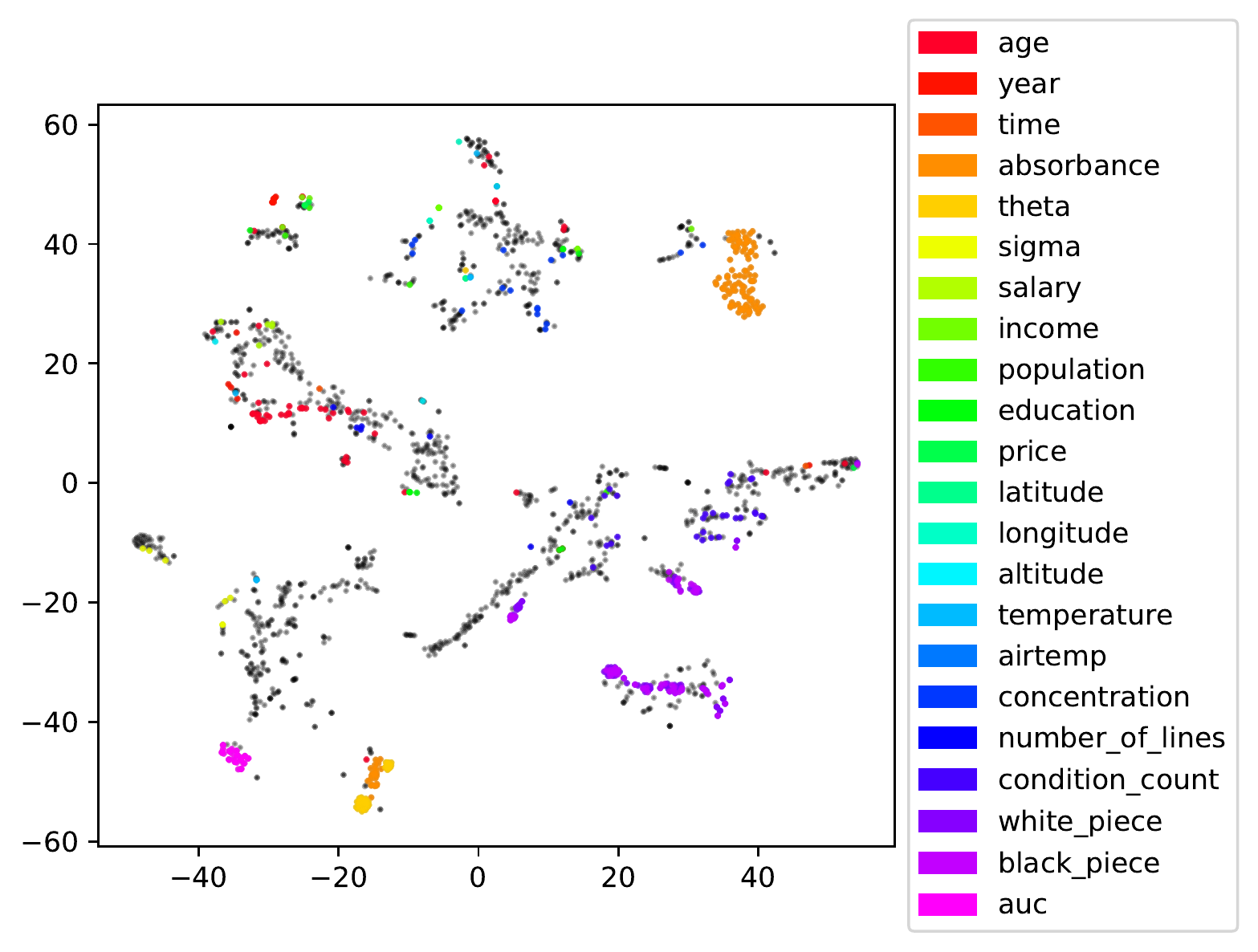}
\vspace*{-2pt}
\caption{t-SNE visualization of our embeddings for 1K numeric OpenML datasets held-out during training. Datasets whose annotated variable lies amongst the most frequently-occurring annotations listed at the right are colored accordingly.  
}
\vspace*{-8pt}
\label{fig:tsne} 
\end{figure}

A major practical issue is the lack of informative variable-names (i.e.\ column labels) in most data tables, and a machine-readable description or codebook for the columns is rarely provided. \cite{Kang03, Chen18, Yi18, Jaiswal13}.
It also is unlikely for different groups producing different datasets to adopt precisely the same nomenclature, even when they are making measurements of the same type.  
For example, the age of a person is recorded as \texttt{Age} in some datasets and \texttt{Years} in others.  
To ensure robustness against idiosyncratically chosen variable-names, the methodology presented here aims to match variables solely on the basis of their data values. 
In settings where adequate column-annotations are available, our methods can nonetheless be straightforwardly combined with standard information retrieval techniques (e.g.\ based on word-embedding of the column label) to leverage the information across both the variable-names and the values of their measurements (c.f.\ \cite{Doan01,dsl,tableunion}).  

The main idea of our approach is to use this dataset repository to learn a vector embedding for each set of data values, such that datasets which stem from the same variable are embedded closer together than those which stem from different variables.  
When trained over a huge and representative data repository, our methodology can crucially adapt these embeddings to the different types of variation that naturally occur amongst real-world datasets.
By pre-storing the embedding vectors for each dataset in our repository, the variables most likely to match a given new dataset can be efficiently identified via fast approximate nearest-neighbor algorithms \cite{fastneighbors}.

\section{Related Work}

\cite{neuralstat, multivae} also recently proposed methods for learning vector representations of datasets, but explore their use in generative modeling rather than our task.
\cite{Chen18,supervisedsemanticlabel,Valera17} consider the more related problem of classifying data from an unknown variable into a predetermined set of meaningful categories such as semantic labels.
Such a multiclass classification setup may however undesirably restrict the number of categories that can be considered and requires careful manual curation.
In practice, there exists an extremely long-tail of different variable types encountered in natural data (and the number of types is even continuously growing over time), which thus is not easily reduced into a limited set of classes.  It is also the case that none of the previously-recorded types of variables may apply for some new datasets.

Given just a single pair of datasets, inference of columns that stem from the same variable is also known as instance-based schema matching 
\cite{Doan01, schemablog, Kang03, Jaiswal13}.  Also related are the tasks of semantic labeling \cite{semtyper, dsl}, dataset search \cite{tableunion}, and Bayesian sets \cite{bayessets}.
However, existing methods from these communities rely on manually-derived features of the data-distribution chosen by authors, unlike the neural network embeddings we employ, which automatically learn features that are naturally important across myriad datasets.  
Unlike our approach, existing tools entirely rely on the stringent assumption that different datasets from the same variable will always be drawn from identical distributions.
As they do not leverage our fixed-size vector representations of datasets, these existing methods also cannot directly exploit fast approximate nearest neighbor methods and instead resort to knowledge-based blocking heuristics to remain computationally viable for a large number of high-dimensional datasets \cite{schemablog, schemamatchingsurvey}.

\section{Methods} 
\label{sec:methods}

To formalize our task, suppose we have a repository $\mathcal{R}$ containing $N$ datasets ${\mathcal{D}_1,\dots,\mathcal{D}_N}$, where each dataset $\mathcal{D}_i$ stems from a single variable $\varletter_i$.  Here, we assume each individual data instance value $x \in \mathcal{D}_i$ belongs to a common space $\mathcal{X}$ shared across datasets, and the instance-values within each $\mathcal{D}_i$ are IID (e.g.\ not time-series).  In this work, we consider the case where $\mathcal{X}$ is either the set of numeric scalar values or the set of (finite-length) strings.
This means that formerly structured data tables have been split in $\mathcal{R}$ into separate datasets comprised of their columns, which is done to ensure we can accurately recognize each individual column within a newly acquired data table.  It is also assumed that some of the datasets stem from the same type of variable, but do not contain the same set of observations ($\varletter_i = \varletter_j, \mathcal{D}_i \neq \mathcal{D}_j$ for some ${1 \le i \neq j \le N}$).
Given a previously-unseen query dataset $\querydata$ generated from an unknown variable at test-time, our system seeks to efficiently infer which of ${\mathcal{D}_1,\dots,\mathcal{D}_N}$ likely stems from the same type of variable as $\querydata$.
Note that the query dataset may match more than one or none of the repository datasets.

A standard approach to determine whether two datasets stem from the same variable bases this decision on the similarity between their data distributions, or summary statistics thereof \cite{Jaiswal13,dsl}.
One way to quantify the distance between two distributions is via:
\begin{equation}
d(P_1, P_2) = \left| \left| \mathbb{E}_{P_1} [h(x)] -  \mathbb{E}_{P_2} [h(x)]  \right|\right| 
\label{eq:dist}
\end{equation}
where the feature map $h$ determines how  distributional-differences are reflected in the resulting distance between the summary statistics formed by average feature values.  For example, a histogram representation of the distribution can be encoded by letting different dimensions of $h(x)$ correspond to indicator functions for different bins.
When $h$ corresponds to the (infinite-dimensional) feature map of a universal RKHS, this distance is known as the maximum mean discrepancy (MMD) and can capture \emph{any} difference in distribution  \cite{mmd}.
Even finite-dimensional $h$ can retain this property if we limit the distributions to certain parametric families.

In this work, we instead implement $h$ as a neural network with parameters $\theta$ and compare distributions via the squared Euclidean distance:
\begin{equation}
d_\theta(P_1, P_2) = \left|\left| \mathbb{E}_{P_1} [h_\theta(x)] -  \mathbb{E}_{P_2} [h_\theta(x)]  \right|\right|_2^2
\label{eq:populationdist}
\end{equation}
We can ensure these representations are able to reflect subtle differences in distributions by employing a high-capacity network with high-dimensional outputs.
If dataset $\mathcal{D}_i$ consists of IID samples from distribution $P_i$, then 
we can empirically use the following distance between datasets:
\begin{align*}
    & d_\theta(\mathcal{D}_1, \mathcal{D}_2) = \left| \left|  h_\theta(\mathcal{D}_1)  - h_\theta(\mathcal{D}_2)  \right|\right|_2^2  \\
    & \text{where } h_\theta(\mathcal{D}_i) = \frac{1}{|\mathcal{D}_i|} \hspace*{-1pt} \sum_{x \in \mathcal{D}_i} \hspace*{-2pt} h_\theta(x)
    \numberthis \label{eq:netdist}
\end{align*}
A fixed-size vector representation of a dataset is thus provided by the averaged output of the neural network ${h_\theta: x \rightarrow \mathbb{R}^k}$, as in \emph{deep sets} \cite{deepsets}.
The computational complexity of $d_\theta(\mathcal{D}_1, \mathcal{D}_2)$ is linear in the datasets' sample size, unlike the quadratic-time MMD statistic, for which linear-time approximations come at the cost of a substantial variance increase \cite{mmd,Chwialkowski15}.  
However, this general approach to infer whether two datasets stem from the same variable solely based on their distributional similarity (e.g.\ via two-sample testing) suffers from three problems: 
\begin{enumerate}
\itemsep0.05em 
\item[(1)] inability to ignore natural variation between datasets containing measurements of the same type of variable 
\item[(2)] inability to distinguish different variables whose data distributions happen to be identical
\item[(3)] computational inefficiency 
\end{enumerate}

To grasp the first problem, consider a variable like \emph{temperature}, which is recorded in some datasets as Fahrenheit and others as Celsius, or from different geographical regions across datasets.  Clearly, the temperature measurements appearing in different datasets may not be distributed similarly at all, an issue described as: batch effects, sampling bias, or covariate shift. 
By learning to ignore natural factors of variation present between different repository datasets from certain variables, the dataset embeddings proposed in our method can still recognize new datasets from these same variables even when their data distributions are dissimilar. 

The second problem with matching variables via distributional similarity arises from the fact that two variables which are different may nonetheless have highly similar data distributions.  Consider for example any Boolean attribute which takes values \texttt{yes} or \texttt{no}. 
Two such attributes that happen to share the same proportion  of \texttt{yes} instances will have identical data distributions, even if these attributes share no relation at all.  
By learning from the data repository which sorts of distributions are naturally common between unrelated variables, our proposed methods remain able to distinguish new variables with these same common distributions. 
Finally, if each dataset contains $O(n)$ samples, identifying the best candidates for a query dataset via distributional similarity would generally require at least $O(N n)$ computation. 
This is undesirably slow for a repository containing many large datasets.  
Our method's reliance on fixed-size dataset embeddings enables the use of fast approximate nearest neighbor search to circumvent the inefficiency of exhaustive pairwise comparison.

\subsection{Modeling Dataset Similarity}

Our general strategy is to learn neural networks which infer the likelihood that two datasets stem from a  shared variable via scalable stochastic gradient methods. 
Let $p_{ij} \in [0,1]$ denote the probability that datasets $\mathcal{D}_i$ and $\mathcal{D}_j$ stem from the same attribute.
We directly estimate this match probability via the parameterized form $p_{ij} = \exp (- D_{ij} )$ where: 
\begin{equation}
 D_{ij}  = d_\theta(\mathcal{D}_i, \mathcal{D}_j) + g_\psi(\mathcal{D}_i) + g_\psi(\mathcal{D}_j)
 \label{eq:matchprobdist} 
\end{equation}
Here, $d_\theta$ is the distance between dataset embeddings defined in (\ref{eq:netdist}), and 
$g_\psi \ge 0$ is a scalar output by another deep sets neural network (with parameters $\psi$) to adjust the match probability:
\begin{equation}
g_\psi(\mathcal{D}_i) = \frac{1}{| \mathcal{D}_i | } \sum_{x \in \mathcal{D}_i}  g_\psi (x) \ \ge 0
\label{eq:probadjust}
\end{equation}
To ensure a nonnegative output value, the final activation used in the output layer of this network is the square function $a(z) = z^2$. 
Note that when the adjustment factor $g_\psi = 0$, the match probability between two datasets is solely determined by the distance between their $h_\theta$-embeddings. 
Datasets that likely stem from the same variable share similar embeddings while those unlikely to match are separated in the embedding-space.
However as discussed previously, the embedding distance in (\ref{eq:netdist}) alone will be unable to distinguish between two different variables whose data happen to be identically distributed.  Thus, $g_\psi$ should learn to output large adjustment values for datasets with a distribution that is common amongst different variables in the repository, as this is the only way to produce a low match-probability between such datasets under our model. Figure \ref{fig:numericrepresentation} shows that empirical performance severely degrades without $g_\psi$.

\subsection{Training Procedure}

Training examples presented to our models are of the basic form $(\mathcal{D}_i, \mathcal{D}_j, y_{ij})$
where $y_{ij} \in \{0, 1\}$ is equal to 1 if and only if $\mathcal{D}_i$ and $\mathcal{D}_j$ stem from the same variable ($\varletter_i = \varletter_j$). In this form, we can learn neural network parameters $\theta, \psi$ by optimizing the standard cross-entropy loss for binary classification: 
\begin{equation}
Loss = - y_{ij} \log p_{ij} - (1-y_{ij}) \log (1-p_{ij}) 
\label{eq:crossentropy}
\end{equation}

Since $g_\psi \ge 0$, the measure $D_{ij}$ defined in (\ref{eq:matchprobdist}) 
satisfies all properties of a metric except for the identity of indiscernibles (recall we explicitly do \emph{not} want this property in order to match/distinguish attributes with different/identical data distributions).  
The match probability definition in (\ref{eq:matchprobdist}) implies our overall cross-entropy loss can be equivalently written as: 
\begin{equation}
Loss =  y_{ij} D_{ij} + (1-y_{ij}) \log \left( 1/\big(1 - e^{-D_{ij}}\big) \right)
\label{eq:redefinedloss}
\end{equation}
which is of a similar geometric form to the loss \cite{Chopra05} proposed to ensure various metric learning desiderata such as a positive margin between positive and negative pairs.  
Under this loss, it is more important to ensure that \emph{un}-matched data are separated rather than matched data sharing nearly the same embedding, a critical property to reduce the false positive rate when searching for variables that might match a new dataset.
Networks $h_\theta$ and $g_\psi$ are learned jointly via the Adam stochastic gradient method applied to this loss function  \cite{adam}.


Rather than sampling pairs of datasets from the repository at a time for training the embeddings, we employ triplets $(\mathcal{D}_a, \mathcal{D}_p, \mathcal{D}_n)$ in which the first two datasets stem from the same variable ($y_{ap} = 1$) and the latter two do not ($y_{an} = 0$). Triplet training which pairs the same anchor sample with both a positive and negative match is a commonly utilized strategy in embedding learning \cite{facenet}.
To calculate stochastic gradients for updating network parameters, we simply draw such dataset-triplets at random from our repository $\mathcal{R}$.
Note that this training strategy balances positive/negative ratios, which may not reflect the desired marginal distribution in a particular application. 
Nonetheless, the estimated match probabilities can be easily recalibrated by estimating the overall variable-match rate amongst random dataset pairs in $\mathcal{R}$ \cite{probadjust}.
Balancing schemes like ours are known to often substantially improve performance in applications with limited positive examples \cite{Batista2004}.
We also experimented with alternative metric-based training objectives and more sophisticated sampling strategies pervasive in deep image embedding methods, but found these empirically led to increased training times and less stability without performance gains for our task (we considered the triplet/contrastive/margin losses applied directly to $D_{ij}$ rather than $p_{ij}$ combined with semi-hard negative mining or distance weighted sampling \cite{samplingmatters}).
\cite{Horiguchi2017} report similar instabilities.

\subsection{Scaling to Large Datasets via Subsampling}

Basing our training procedure on stochastic gradient methods ensures it can scale to large repositories that contain many datasets.
However, since each training example is comprised of entire datasets, the time to process these examples via our deep sets network depends linearly on the datasets' sample size, which may be undesirably inefficient.
When $\mathcal{D}_i$ is a large dataset, we instead rely on a smaller subset of randomly selected samples from $\mathcal{D}_i$ to obtain our dataset representations $h_\theta(\mathcal{D}_i), g_\psi(\mathcal{D}_i)$, which are subsequently used to compute the measures $D_{ij}$ in (\ref{eq:matchprobdist}).

By Jensen's inequality, our estimate in (\ref{eq:netdist}) of the population distance between underlying distributions (\ref{eq:populationdist}) is slightly biased upward.  
Our definition employs the V-statistic formulation discussed in \cite{mmd} in the case of MMD, rather than a U-statistic.  Thus, our stochastic gradient method 
effectively optimizes
an upper bound for loss that would be incurred based on the embeddings of the underlying population distributions, where the  tightness of this bound depends on the datasets' sample sizes.
Theorem \ref{thm:hoeffding} below nonetheless ensures that our subsampled dataset embedding distances closely approximate their underlying population counterparts.

\begin{thm} Let $\displaystyle {B = \max_{x \in \mathcal{X}}  \left| \left| h_{\theta}(x) \right|\right|^2 }$.  When $\widetilde{\mathcal{D}}_i, \widetilde{\mathcal{D}}_j$ are subsamples of size $n$ from a given pair of datasets $\mathcal{D}_i, \mathcal{D}_j$:
\begin{equation*} 
\Pr \left( \big| d_\theta(\widetilde{\mathcal{D}}_i, \widetilde{\mathcal{D}}_j) - d_\theta(P_i, P_j) \big| > \epsilon \right) \le 4 \exp \left( - \frac{n \epsilon^2}{32 B^2}\right)
\vspace*{-1em}
\end{equation*}
\label{thm:hoeffding}
\end{thm}
The $d_\theta$ quantities above are defined as in (\ref{eq:populationdist})-(\ref{eq:netdist}), and  
the probability is taken over the sampling of datasets from the underlying distributions as well as the random subsampling procedure.  
Here, we have implicitly assumed both datasets contain at least $n$ instances and our neural network outputs are bounded.  The  proof of this theorem directly follows from a combination of the triangle and Hoeffding inequalities.
A similar concentration bound may be derived for the differences ${ \big| g_\psi(\widetilde{\mathcal{D}}_i) - \mathbb{E}_{P_i} [g_\psi(x) ]  \big| }$. Together, these bounds ensure dataset-subsampled-estimates of $D_{ij}$ and $p_{ij}$ are concentrated around the values they would take if provided with the actual embeddings of the underlying population distributions. 
In our applications, we train our embedding networks by drawing subsamples of 1K instance values (when sample-size exceeds this) at which point these empirical approximations are likely quite accurate.  By batching such subsamples on a GPU, we can compute the vector embedding of a subsampled dataset in a single simultaneous pass through each neural network.  

\subsection{Scarcity of Positive Matches}

In practice, it may be difficult for human annotators to comprehensively examine all pairs of attributes in a large dataset repository.  The number of annotated variable matches may thus be small relative to the possible pairings of datasets (i.e.\ $y_{ij}$ may be missing for many $\mathcal{D}_i, \mathcal{D}_j$ pairs). 
Since it is quite unlikely that a random pair of datasets would stem from the same variable, we can safely assume $v_i \neq v_j$ (i.e.\ $y_{ij} = 0$) for all $i,j$ except for pairs where a variable match has been indicated.  
Ideas from positive and unlabeled learning \cite{pu} might instead be used to recalibrate our predicted match probabilities (for example, leveraging the proportion of matched pairs among all pairs considered by the annotator).  

A limited number of positive matches also limits the quality of our learned embeddings.  
To obtain additional positive examples for supervising our neural network training, we employ the following training data augmentation strategy: A single dataset in the repository is randomly split into two disjoint subsets, which are treated as separate datasets whose variables are to be matched. These pairs are then added as additional positive examples to our training set. We repeat this procedure for randomly selected datasets and split them into randomly sized partitions, such that we can generate as many augmented samples as desired. 
Note that for these augmented samples, a measure of distributional similarity does in fact suffice to determine they should be matched.
Thus, their inclusion in the training set encourages our embedding distance to behave like a standard two-sample statistic for data from unique variables without any observed match.

\subsection{Neural Network Models for Different Data Types} 
\label{sec:networkspecifics}
\begin{figure*}[tb] \centering
\textbf{(A)} \hspace*{220pt} \textbf{(B)}  \\[-0.05em]
\includegraphics[width=0.4\textwidth]{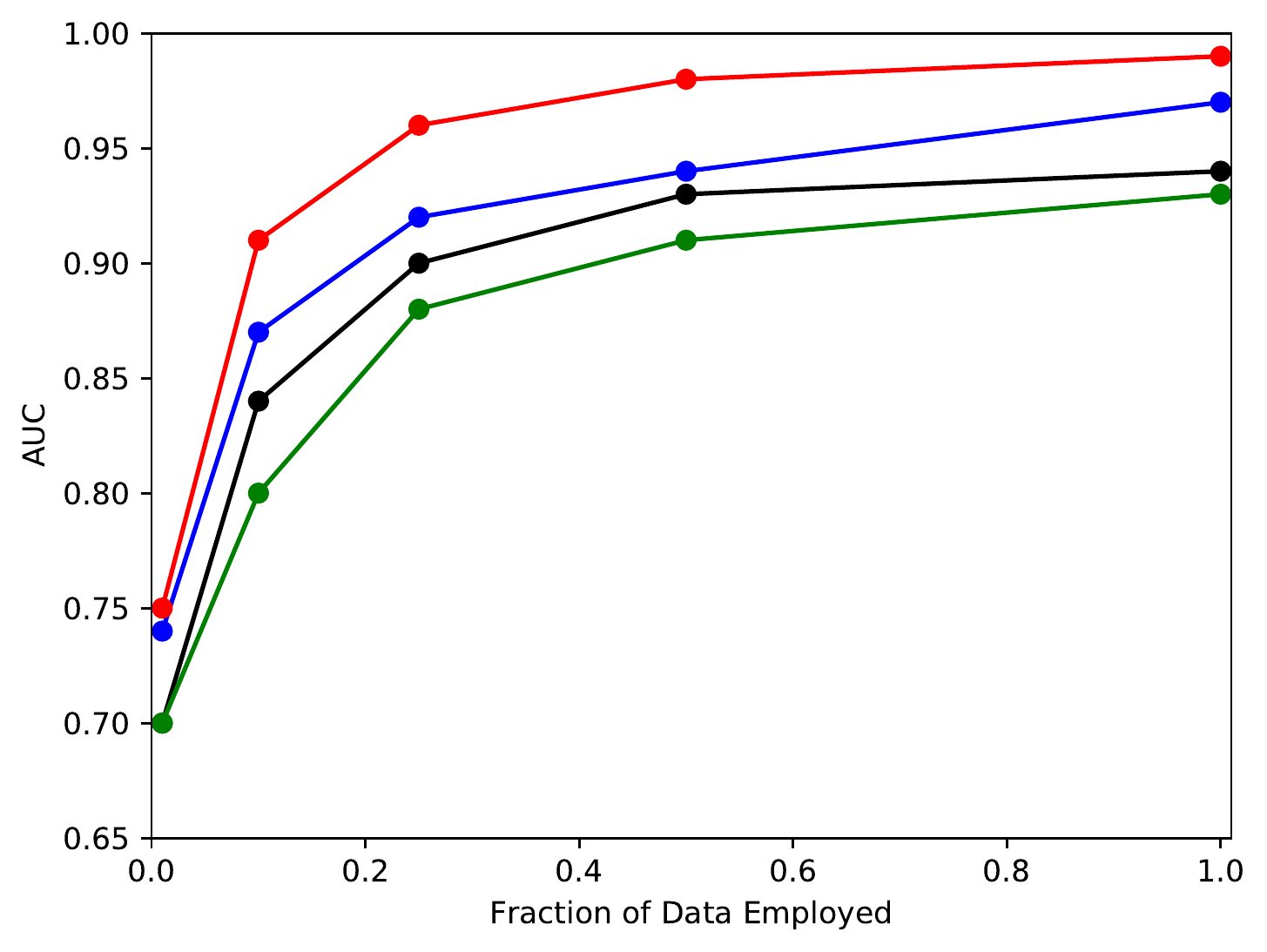}
\hspace*{0.05\textwidth}
\includegraphics[width=0.4\textwidth]{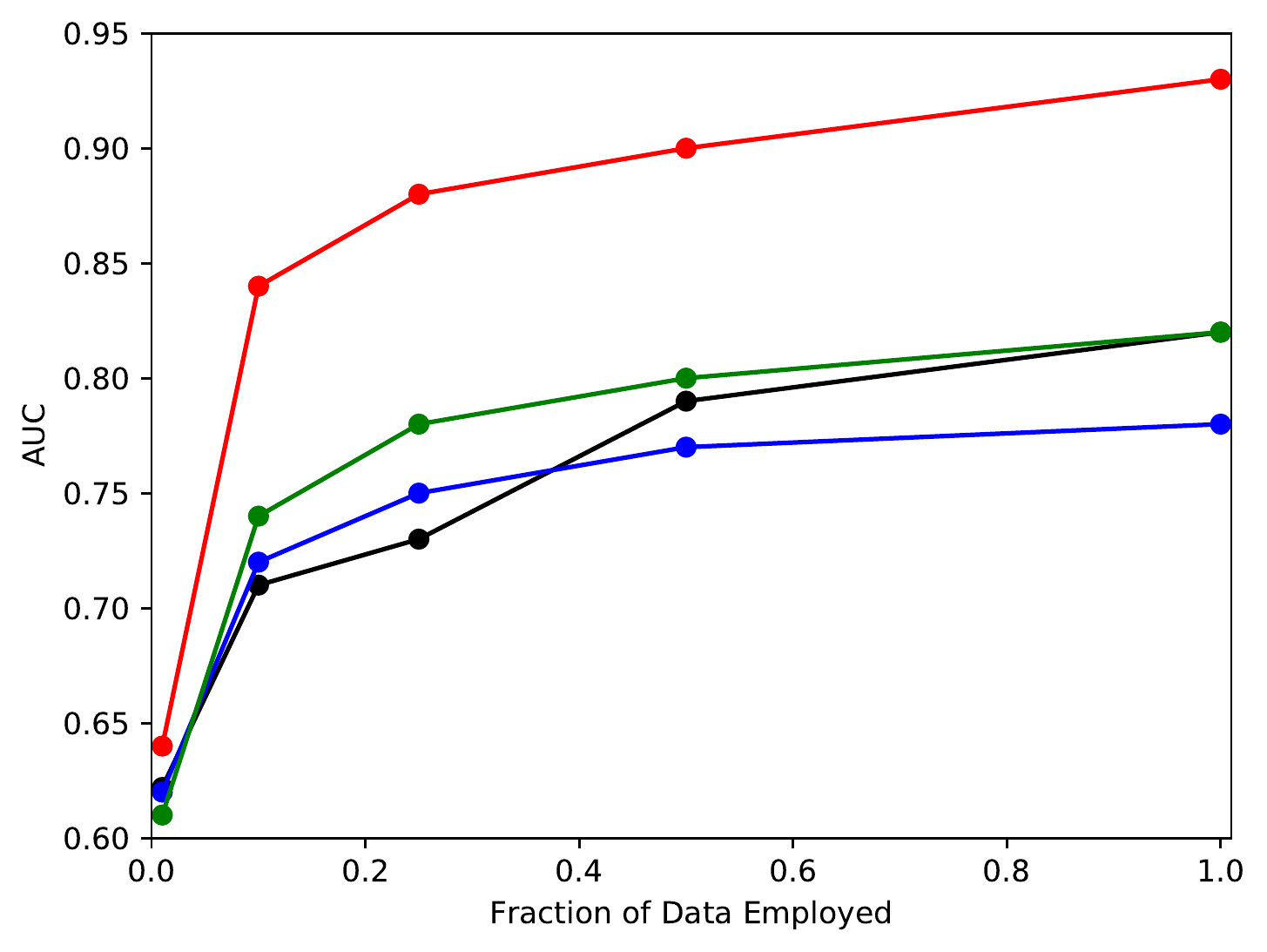}
\caption{Match/no-match classification performance for variations of our \emph{Embed-Num} method applied to various-sized subsamples of numeric datasets under:  \textbf{(A)} \emph{Split} and \textbf{(B)} \emph{Diff} settings.  
Variants use embedding networks that operate upon different numeric input representations: original  floating point values (black), two-dimensional exponent-mantissa vector encoding (blue), and 32-bit binary representation discussed in \S\ref{sec:networkspecifics} (red).  In green, we show ablation results for \emph{Embed-Num} operating on same 32-bit binary input representation, but with probability adjustment network removed (i.e.\ $g_\psi = 0$).
}
\label{fig:numericrepresentation} 
\end{figure*}

Throughout this work, 300-dimensional vector embeddings are used as the output of $h_\theta$, regardless of the data type.
For embedding numeric data where $x \in \mathbb{R}$, both $h_\theta$ and $g_\psi$ are three layer fully-connected networks with 300 units in each hidden layer, ReLU activations in  $g_\psi$, and tanh activations in $h_\theta$ (including for the output layer).
When our repository contains a diverse collection of variables, we may need to embed numeric data across an arbitrary input range (rescaling continuous datasets is undesirable as crucial variable information is encoded in the magnitude of their values).  Unfortunately, standard neural networks perform poorly across inputs of widely-differing magnitudes \cite{nalu}.
Thus, rather than taking as input the raw floating-point value corresponding to each numeric data instance $x$, $h_\theta$ and $g_\psi$ take as input a 32 dimensional vector of binary features encoding the 32-bit signed binary representation of $x$ instead.  
This binary input representation enables the network to more stably operate over diverse input ranges  in different datasets, and consistently results in improved performance (Figure \ref{fig:numericrepresentation}).

While supervised learning pipelines typically bin data with string fields into categories encoded as one-hot vectors, the original string values and their distribution contain important information about the data-generating variable.  Our methods instead operate directly on these strings, producing a vector representation of each string that is subsequently pooled in our deep sets embedding.  
Note that when the number of unique strings in a dataset is far less than the sample size, our deep set embeddings can be efficiently computed as a multinomially-weighted sum of string embeddings.  
Two types of string data may be encountered in practice: Datasets whose entries come from natural language (e.g. names and other standard text fields), and datasets whose values are general strings (e.g.\ phone-numbers or domain-specific vocabulary).  
In order to embed datasets of the former type, we compute the 300-dimensional \texttt{fastText} vector representation\footnote{Pretrained word vectors from \url{http://fastText.cc/}} of each string value \cite{fasttext}, which is then fed as input to fully-connected deep sets networks $h_\theta, g_\psi$ consisting of the same architecture as described previously.  
This use of pretrained word vectors ensures that known semantics of the values are taken into consideration when comparing datasets (e.g.\ that \texttt{US} and \texttt{America} refer to the same entity). 
For a particular string dataset, we determine whether our \texttt{fastText} embedding method is appropriate or not by ascertaining whether a majority of the strings in each data field entail words present in the \texttt{fastText} vocabulary.  When one data field contains multiple words, we simply average their \texttt{fastText} vectors.

For embedding other datasets where each field $x$ is simply an arbitrary finite-length string not from a known language, both $h_\theta, g_\psi$ are taken to be bidirectional character-level Long Short-term Memory (LSTM) recurrent networks \cite{lstm}, which employ two stacked layers with 128 hidden units in each direction.  We concatenate the final hidden states in each direction to form a vector representation of the entire string.  This same string-representation is then passed through separate fully connected layers in $h_\theta$ and $g_\psi$; the former outputs a 300-dimensional embedding with $tanh$ activation, the latter a single nonnegative scalar value.

\subsection{Approximate nearest neighbor search}
\label{sec:fastneighbor}

Given a query dataset $\querydata$, we wish to quickly identify the repository datasets most likely to stem from the same variable, which correspond to those $j \in \{ 1,\dots,N \}$ for which $D_{\querychar{}j}$ is the smallest.
Recalling the definition in (\ref{eq:matchprobdist}), $D_{\querychar{}j}$ can be expressed as a squared  Euclidean distance between:
$$\Big[ h_\theta(\mathcal{D}_j) ; \sqrt{g_\psi(\mathcal{D}_j)} ; 0 \Big]  \ \text{ and } \ \Big[ h_\theta(\querydata) ; 0 ; \sqrt{g_\psi(\querydata)} \Big]
$$
where the final dimensions of these vectors are simply two numbers appended to the dataset embedding vector:   the scalar factor produced by our probability-adjustment network and a zero value.  
For each repository dataset $\mathcal{D}_j$, we compute $[ h_\theta(\mathcal{D}_j) ; \sqrt{g_\psi(\mathcal{D}_j)} ; 0 ]$ and pre-store it.
At test-time, $[ h_\theta(\querydata) ; 0 ; \sqrt{g_\psi(\querydata)} ]$ can be efficiently computed via a forward pass through our deep sets networks.

Having re-expressed our similarity function as a simple distance between vectors that each individually depend only on one dataset, we can leverage approximate nearest neighbor algorithms to efficiently find the most promising candidates for a query dataset. Specifically, we employ the approach of \cite{fastneighbors} which is known to be fast and accurate in practice.
Note that it is our particular formulation of the match probability in (\ref{eq:matchprobdist}) which allows the learned dataset similarity to be expressed as Euclidean distance between dataset-specific embedding vectors.  Approximate nearest neighbor methods are not as easily leveraged for a general neural network architecture that jointly operates on both datasets.

\section{Experiments}
\label{sec:experiments}


Our experiments are based around a publicly-available large-scale source of tabular data stored in the OpenML repository\footnote{\url{http://www.openml.org/}} \cite{openml}.  
OpenML serves as a nice data repository as it is not only well-organized (with the performance of many learning algorithms stored for each dataset), but is also continually growing over time as users upload new data to the platform.
Each column from a data table is split into a separate dataset, within which we omit entries with missing values for simplicity.  
We consider all data files pertaining to either regression or classification tasks, which appeared to be the best organized data.
We partition the OpenML data into disjoint groups of numeric, language and general-string datasets which are handled separately.
This means we only consider matching numeric datasets with numeric datasets, language data with other language data, and strings with strings, as well as training separate embedding models for each  data type.

To programmatically identify matched variables across different columns of OpenML data tables (i.e.\ what we refer to as a \emph{dataset} in this work), we leverage the provided column labels and employ standard techniques from schema-matching.  Namely, if two column labels (or tokenized versions thereof) are present in the set of pretrained \texttt{fastText} word vectors, we consider whether the cosine similarity between their vector representations exceeds a highly conservative threshold (0.9).  If the column label strings cannot be tokenized into valid words with corresponding pretrained word vectors, then we instead consider whether the Jaro-Winkler string distance between them exceeds a separate conservative threshold (0.95), as proposed in \cite{jaromatch}.  
While we rely on the provided column labels to identify true variable matches in the OpenML data, none of these labels are accessible to the different learning methods studied in our evaluations.
Note that our strategy to identify variable matches based on provided names ensures columns with identical names are always matched.  As this is not always desirable, we manually assemble a list of common but vague column names comprised of generic descriptors (such as: \texttt{attribute}, \texttt{variable}, \texttt{column}, as well as names with under 3 characters) and remove data corresponding to columns with these uninformative labels from our analysis.

The data tables of each type are separated into two groups: one whose columns form the datasets contained within a training repository $\mathcal{R}$ (to represent previously-observed datasets used for training our embeddings) and another whose columns form a collection of held-out datasets $\mathcal{T}$ only used during evaluations (analogous to the ``test set'' in supervised learning).  For a small subset of training datasets in $\mathcal{R}$, we partition their observations into two halves, one of which is removed from $\mathcal{R}$ and
added as a separate dataset to another collection $\mathcal{S}$ also held-out for evaluations.  Table \ref{tab:datasummary} summarizes the OpenML data.
We simply train our embedding networks on each repository $\mathcal{R}$ as described in Section \ref{sec:methods}, until the training loss in (\ref{eq:crossentropy}) no longer consistently improves (we did not note  overfitting). 
Once our embedding models have been trained on the appropriate OpenML repository $\mathcal{R}$, we perform all evaluations in this paper with no additional training of these models. 
While our methods operate at the level of individual columns (what we refer to as a dataset in this work), we ensure that our analysis always compares data from different randomly-drawn OpenML tables to reflect realistic use-cases (and avoid overrepresentation of high-dimensional tables).

\begin{table*}[t]
\caption{Summary of the OpenML data repositories used in this work.   
For each partition of the various  data types, we list the: 
number of original data tables, 
number of datasets ($N$) created from the columns of these tables, 
sample size of these datasets (mean $\pm$ standard deviation). 
The second column lists the number of matched pairs between OpenML datasets of each type,  identified based on the provided variable names.
}
\label{tab:datasummary}
\begin{center}
\begin{small}
\begin{sc}
\begin{tabular}{lccccc}
\toprule
Data type & Num.\ matches & Partition & Num.\ tables & Num.\ datasets & Sample sizes \\
\midrule
& & $\mathcal{R}$  & 616 & $1.0e6$ & $490 \pm 2e3$ \\
Numeric & $1.6e4$   & $\mathcal{T}$  &  100 & $4.3e4$ & $3.1e3 \pm 8e3$   \\
&  & $\mathcal{S}$    & - & 800 & $3.2e4 \pm 7e3$   \\
\midrule
&  & $\mathcal{R}$    & 308 & $2.7e3$ & $1.2e3 \pm 5e3$ \\ 
Language & $3.6e3$  &  $\mathcal{T}$  & 50 & 910 & $2.5e3 \pm 7e3$ \\ 
&  & $\mathcal{S}$    & - & 200 & $1.8e3 \pm 7e3$ \\
\midrule
&  & $\mathcal{R}$    & 185 & $1e3$ & $810 \pm 5e3$ \\
String & $3.0e3$ & $\mathcal{T}$  & 50 & 180 & $560 \pm 2e3$ \\
&  & $\mathcal{S}$    & - & 100 & $430 \pm 3e3$ \\
\bottomrule
\end{tabular}
\end{sc}
\end{small}
\end{center}
\end{table*}

\subsection{Alternative Methods}

When presenting results, we refer to our methods as \emph{\textbf{Embed-Num}} when applied to numeric datasets, \emph{\textbf{Embed-Txt}} when applied to string datasets with text fields represented via \texttt{fastText} vectors, and \emph{\textbf{Embed-Str}} when applied to general string data.
For numeric data, we compare against alternative methods where the $D_{ij}$ used to determine match probabilities between datasets are instead given by: 

\noindent \emph{\textbf{MeanSD}}:  Empirical mean difference plus the empirical standard deviation difference between two datasets.

\noindent \emph{\textbf{KS}}: 1 minus the $p$-value of the Kolmogorov-Smirnov test for an arbitrary difference between distributions \cite{ks}.

\noindent \emph{\textbf{MMD}}: Maximum Mean Discrepancy with RBF kernel. For computational efficiency, we use the linear-time estimator \cite{mmd}. 

\noindent \emph{\textbf{SCF}}: More powerful linear-time MMD estimator from \cite{Chwialkowski15}.
We use the Smooth CF variant with $J = 10$ test frequencies.

For string data, we compare against the following alternate measures of dataset-distance $D_{ij}$ (the latter methods are suited for natural language data, while \emph{Jaccard} is more appropriate for arbitrary strings with uniformative \texttt{fastText} vectors):

\noindent \emph{\textbf{Jaccard}}: Jaccard distance between sets of unique string-values appearing in instances within each dataset, as considered by many schema-matching systems \cite{Rahm11,automatch,schemablog}.

\noindent \emph{\textbf{mWordVec}}: Distance between means of the  \texttt{fastText} word vectors computed from the entries in each dataset.

\noindent \emph{\textbf{pWordVec}}: 1 minus $p$-value of Hotelling $T^2$ statistic for the difference in the above mean \texttt{fastText} vectors, used in \cite{tableunion}.

\section{Results}  
\label{sec:results}

Using t-SNE to visualize our dataset embeddings  (Figure \ref{fig:tsne}) reveals that these vectors capture quite a bit of information about the  annotated variable names, despite never having previously encountered these data or any variable-name annotations at all. 
Our first quantitative evaluation involves the following binary classification problem: given a random pair of held-out datasets, determine whether or not their variables match.  We consider balanced instances of this problem (with same number of matched and unmatched test pairs), where performance is quantified via the area under the ROC curve (AUC). 
To evaluate across different sample-size regimes, we also apply our methods to down-sampled fractions of each test dataset. 
Two variants of this problem are considered:

\noindent \emph{\textbf{Split}}: the positive class of matched examples consists of pairs formed by partitioning the 
observations from a single test dataset in $\mathcal{T}$ into two equal-sized datasets.

\noindent \emph{\textbf{Diff}}: the positive class of matched examples consists of pairs that are two different test  datasets from $\mathcal{T}$ whose variable names indicate a match.

\begin{figure*}[tb] \centering
\textbf{(A)} \hspace*{220pt} \textbf{(B)}  \\[-0.1em]
\includegraphics[width=0.4\textwidth]{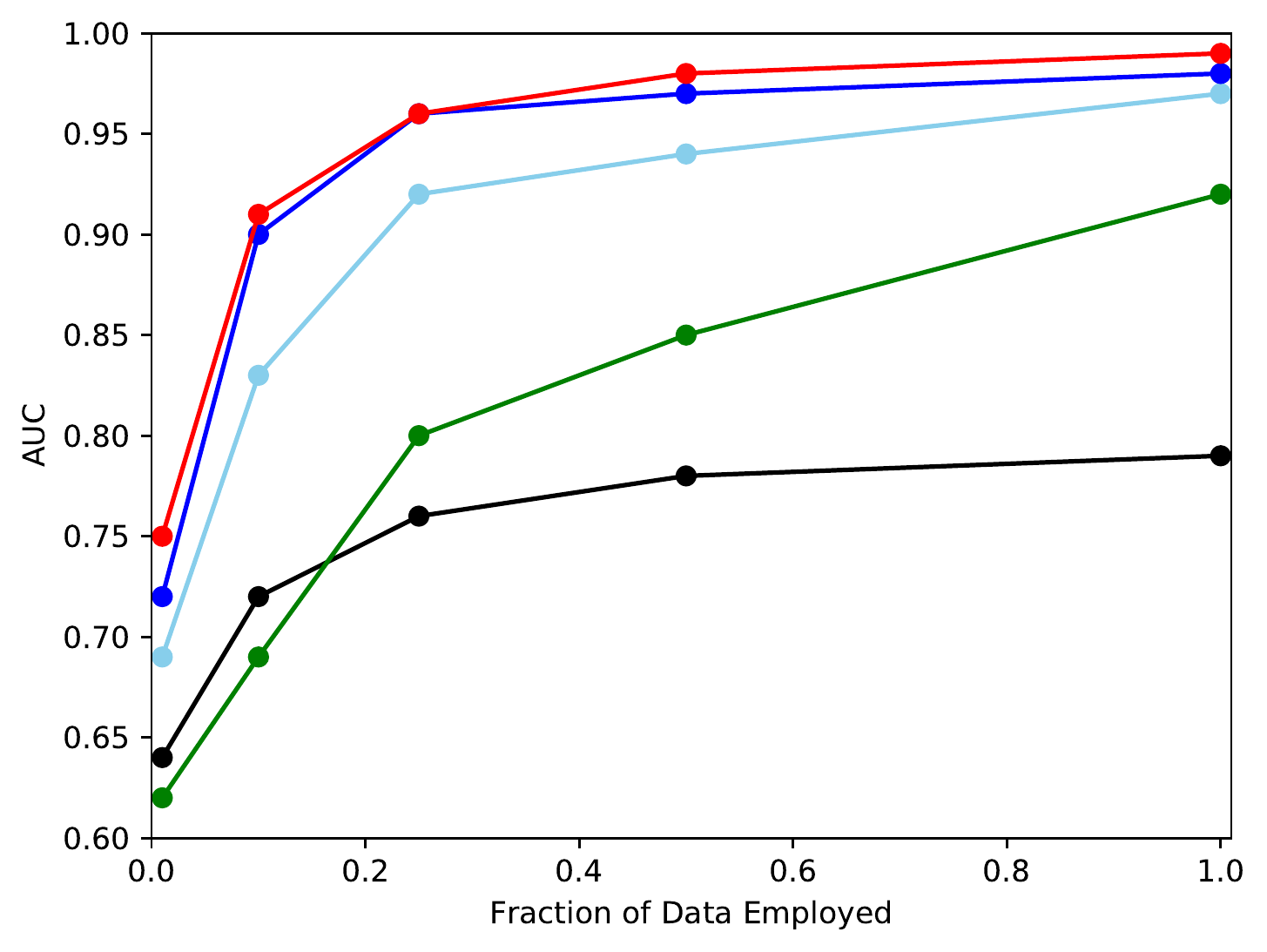}
\hspace*{0.05\textwidth}
\includegraphics[width=0.4\textwidth]{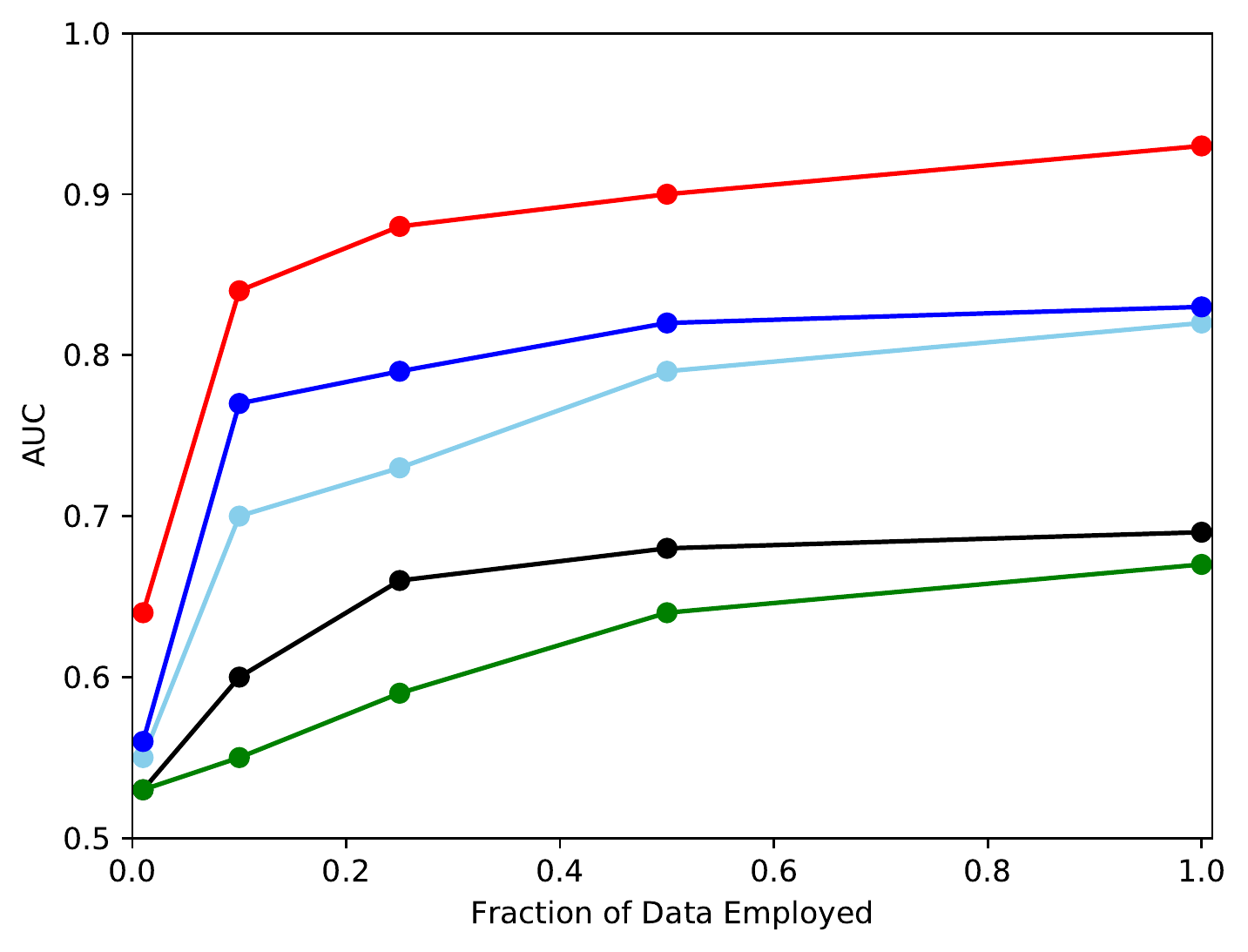}
\vspace*{-2pt}
\caption{Match/no-match classification performance of: \emph{MeanSD} (black), \emph{KS} (green), \emph{MMD} (light blue), \emph{SCF} (blue), and \emph{Embed-Num} (red) estimators on various-sized subsamples of numeric datasets under:  \textbf{(A)} \emph{Split} and \textbf{(B)} \emph{Diff} settings.
}
\label{fig:aucnumeric} 
\end{figure*}

\begin{figure*}[tb] \centering
\textbf{(A)} Language data (\emph{Split}) \hspace*{135pt} \textbf{(B)} Language data (\emph{Diff}) \\[0em]
\includegraphics[width=0.4\textwidth]{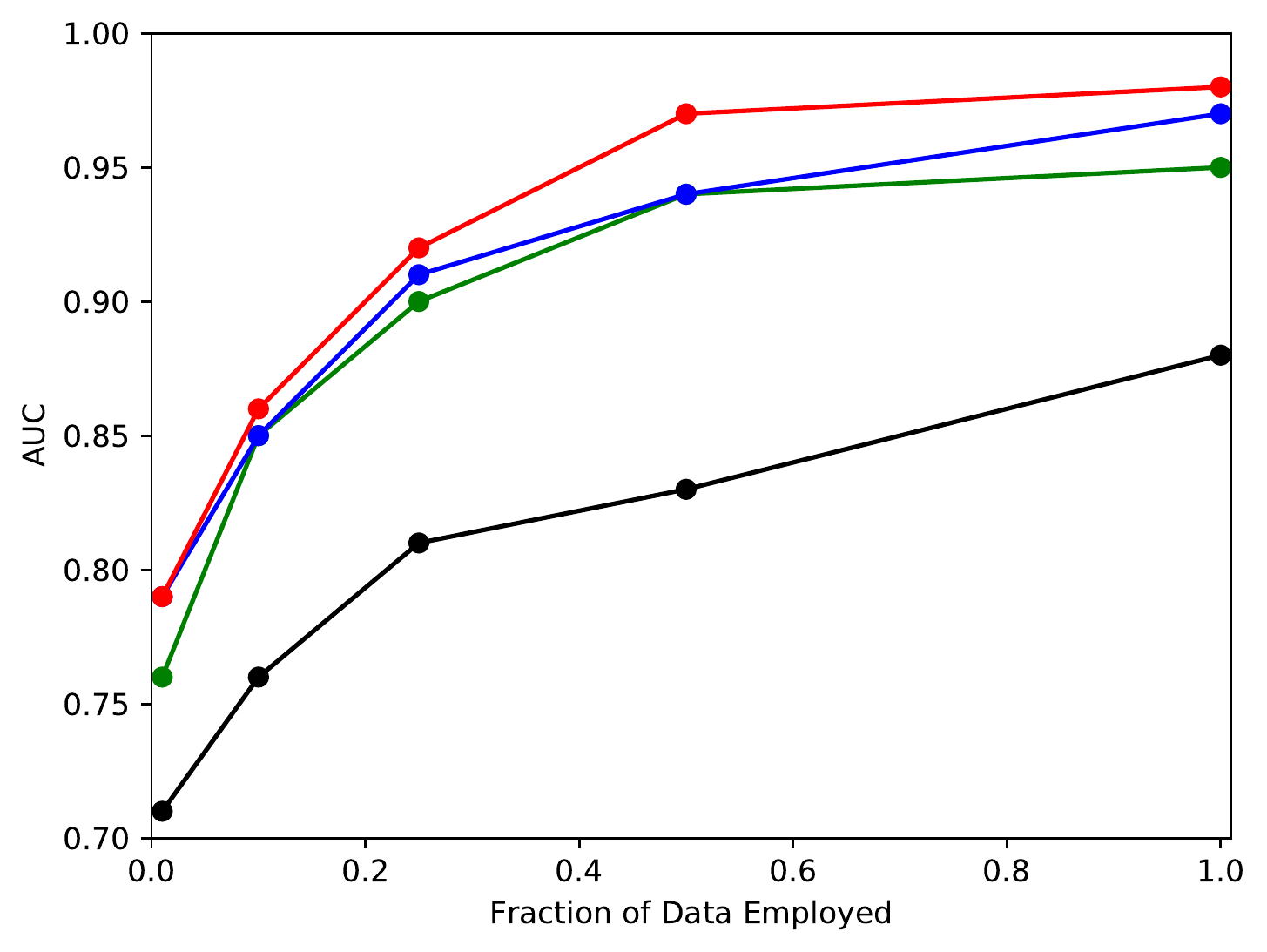}
\hspace*{0.05\textwidth}
\includegraphics[width=0.4\textwidth]{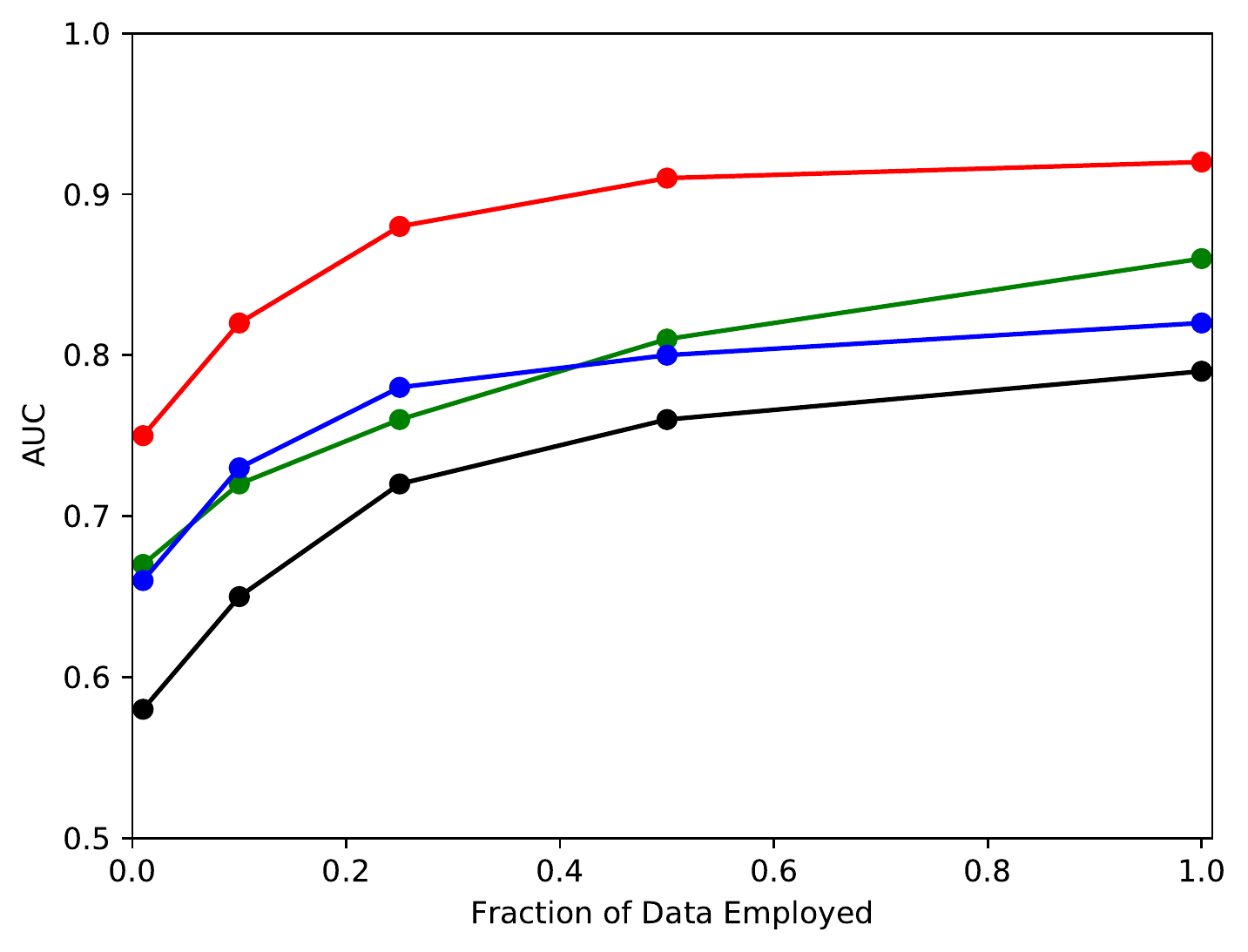} \\[0.2em]
\textbf{(C)} String data (\emph{Split}) \hspace*{150pt} \textbf{(D)} String data (\emph{Diff})  \\[0.0em]
\includegraphics[width=0.4\textwidth]{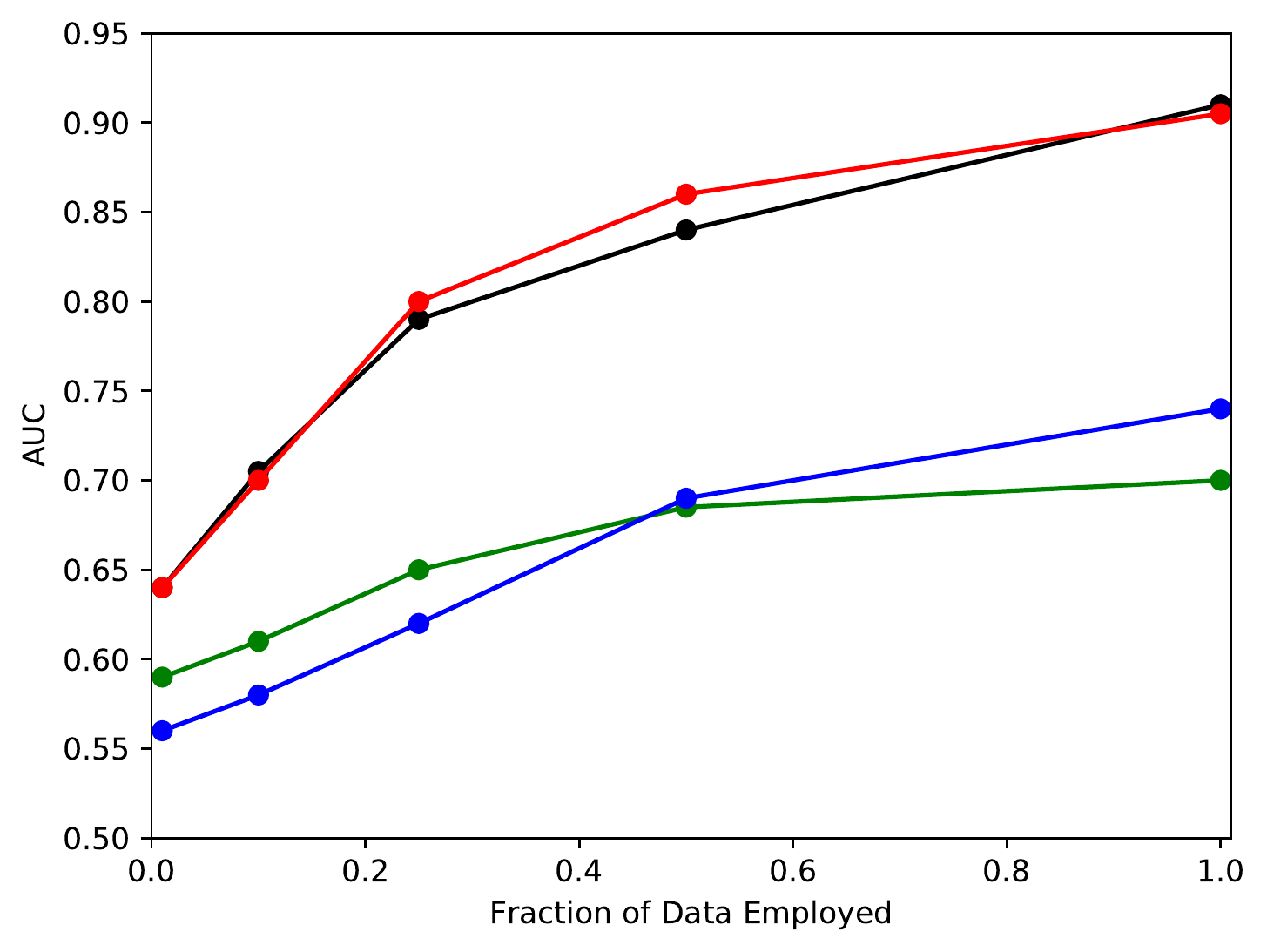}
\hspace*{0.05\textwidth}
\includegraphics[width=0.4\textwidth]{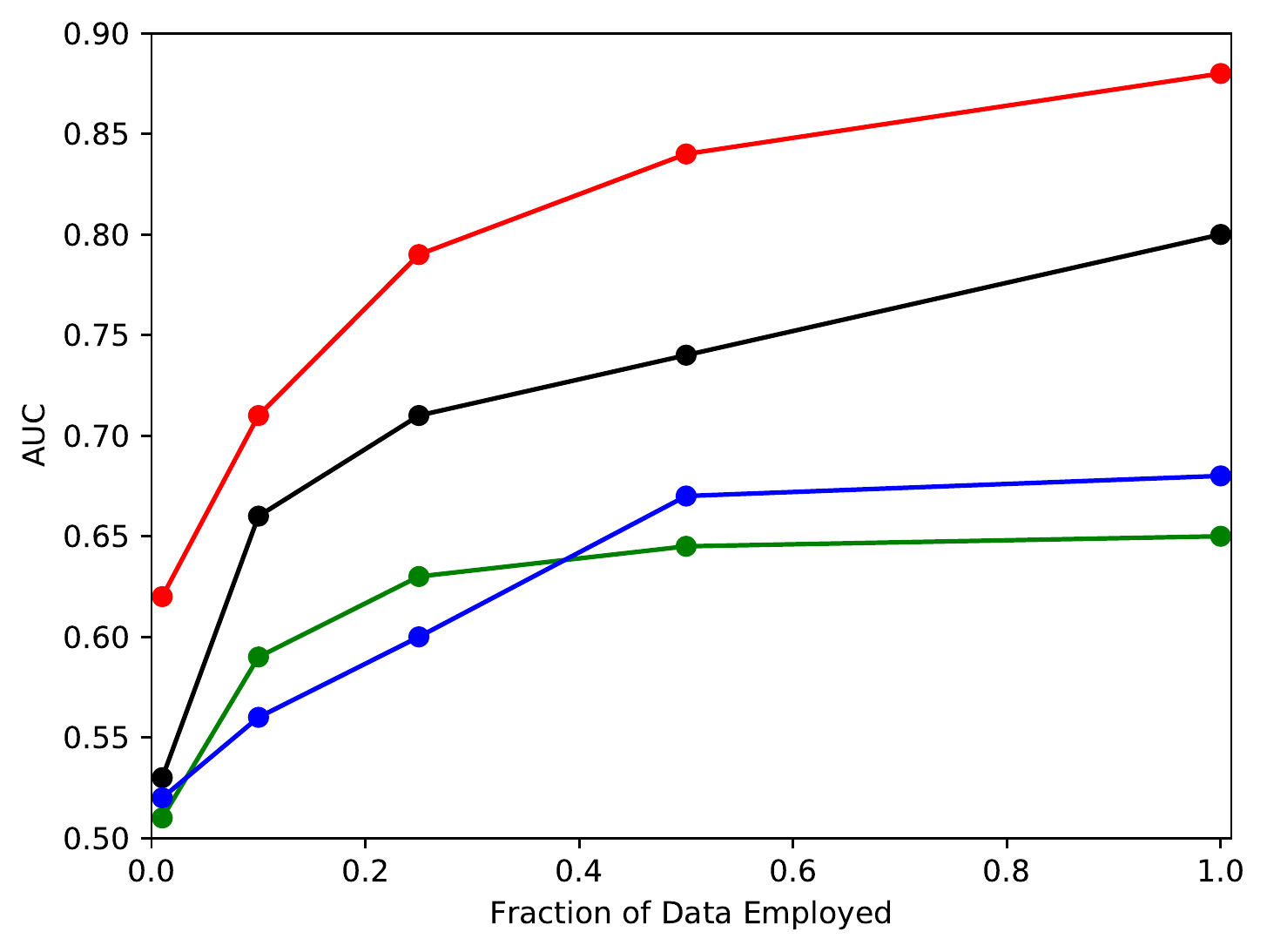}
\caption{Match/no-match classification performance of: \emph{Jaccard} (black), \emph{mWordVec} (green), \emph{pWordVec} (blue), and our methods (red): \emph{Embed-Txt} in \textbf{(A)}-\textbf{(B)}, \emph{Embed-Str} in \textbf{(C)}-\textbf{(D)}, in subsampled held-out language/string datasets from \emph{Split}/\emph{Diff}.
}
\label{fig:auclanguagestring} 
\end{figure*}

In both variants, the negative class of unmatched examples consists of random pairs of test datasets in $\mathcal{T}$ whose variable names do not match.
Note that \emph{Split} is a simpler problem which can be solved by accurate measures of statistical difference between distributions (without learning to ignore natural variation between different datasets that stem from the same type of variable).
Figure \ref{fig:aucnumeric} depicts the classification results for 1000 test pairs sampled from each class.  In the \emph{Diff} setting, none of the alternative methods (which are based solely on distributional-similarity) performs nearly as well as our approach, validating the reasons listed in \S\ref{sec:intro} and \S\ref{sec:methods}. Even in the \emph{Split} setting where the alternative methods are appropriate, our approach is able to slightly outperform standard two-sample methods having learned to adapt its embeddings to types of numeric data that appear in practice. 
Figure \ref{fig:auclanguagestring} shows analogous results for the language/string data.

Next, we consider the task of retrieving repository datasets whose variables match that of a given (held-out) query dataset. Repository datasets are ranked based on their match probability with the query (monotonically-related to measure of dataset distance used), and we report the (average) number of correct variable matches present among the top-$k$ ranked results for each query (more appropriate than precision/recall in our setting where there are so few variable matches).
Two variants of this problem are considered:

\noindent \emph{\textbf{Split-Retrieve}}: the query dataset comes from $\mathcal{S}$ and matches exactly one dataset in $\mathcal{R}$ (as both datasets originate from the same column which has been partitioned in two).

\noindent  \emph{\textbf{Diff-Retrieve}}: the query dataset comes from  $\mathcal{T}$ and matches all datasets in $\mathcal{R}$ with similar variable names.  We try  100 query datasets randomly chosen from a subset of the datasets in $\mathcal{T}$ for which we know at least one such match in $\mathcal{R}$ exists.  

\emph{Split-Retrieve} is again a simpler problem that can in theory be solved by accurate two-sample measures of distributional differences.
In Table \ref{tab:retrievenumeric}, the best performance values for each variant are highlighted in bold, indicating that our approach tends to have higher precision in these dataset retrieval tasks.  
For example, when our query consists of numeric values from a column named \texttt{age} (part of a table annotated as survival data), the top 3 matches returned by \emph{Embed-Num} are all columns with the same label \texttt{age}, belonging to tables respectively annotated as audiology, diabetes, and breast tumor data. Our embeddings correctly recognize that these four datasets all correspond to measurements of patient ages in medical studies.

\begin{table}[t!] 
\caption{Number of correct matches @$k$ for retrieving  datasets from $\mathcal{R}$ in \emph{Split-Retrieve}  (unshaded) and \emph{Diff-Retrieve} (shaded) settings 
(averaged over 100 query datasets).  Results are separated for OpenML datasets of different data types: numeric (top rows), language (middle rows), string (bottom  rows).  The final column depicts the \emph{calibrated-recall} (see \S\ref{sec:results}).
}
\label{tab:retrievenumeric}
\vspace*{-0.8em}
\begin{center}
\begin{small}
\begin{tabular}{llllll}
\toprule
& \multicolumn{1}{l}{ \begin{sc} Method \end{sc} } 
& \multicolumn{1}{l}{ $k=1$ } & \multicolumn{1}{l}{ $k=5$ } & \multicolumn{1}{l}{ $k = 10$ } & \multicolumn{1}{l}{ Recall } \\
\midrule
& \emph{MeanSD} & 0.4 & 0.51 & 0.58  & -  \\
& \emph{KS} & 0.46 & 0.59 & 0.67 & -  \\
& \emph{MMD} & 0.48 & 0.62 &  0.69  & - \\
& \emph{SCF} & \textbf{0.49} &  0.65 & 0.72 & - \\
& \emph{Embed-Num} & 0.48 & \textbf{0.66} &  \textbf{0.74} & -  \\
\rowcolor[gray]{.92} & \emph{MeanSD} & 0.35 & 0.44 & 0.53  & 0.52    \\
\rowcolor[gray]{.92} & \emph{KS} & 0.36 & 0.46 & 0.59 & 0.6  \\
\rowcolor[gray]{.92} & \emph{MMD} & 0.33 & 0.45 &  0.52  & 0.6  \\
\rowcolor[gray]{.92} & \emph{SCF} & 0.33 &  0.4 & 0.55 & 0.62   \\
\rowcolor[gray]{.92} & \emph{Embed-Num} & \textbf{0.42} &   \textbf{0.61} & \textbf{0.67} & \textbf{0.71}  \\ 
\midrule 
\multirow{-13}{*}{\STAB{\rotatebox[origin=c]{90}{ {\normalsize  Numeric data}  }}} &
\emph{Jaccard} & 0.41 & 0.56 & 0.62   & - \\
& \emph{mWordVec} & 0.49 & 0.59 & 0.68  & -  \\
& \emph{pWordVec} & 0.51 & 0.63 &  0.7  & - \\
& \emph{Embed-Num} & \textbf{0.54} & \textbf{0.66} &  \textbf{0.71} & -  \\
\rowcolor[gray]{.92} & \emph{Jaccard} & 0.32 & 0.38 &  0.52  & 0.59  \\
\rowcolor[gray]{.92} & \emph{mWordVec} & 0.4 & 0.46 &  0.59  & 0.67  \\
\rowcolor[gray]{.92} & \emph{pWordVec}  & 0.43 &  0.49  & 0.59 & 0.68 \\
\rowcolor[gray]{.92} & \emph{Embed-Num} & \textbf{0.49} &   \textbf{0.59} & \textbf{0.67} & \textbf{0.75}  \\ 
\midrule
\multirow{-11}{*}{\STAB{\rotatebox[origin=c]{90}{ {\normalsize  Language data}  }}} & 
\emph{Jaccard} & 0.44 & \textbf{0.67} &  0.7   & -  \\
& \emph{mWordVec} & 0.31 & 0.38 &  0.6  &  -  \\
& \emph{pWordVec} & 0.34 & 0.4 &  0.62  & - \\
& \emph{Embed-Num} & 0.44 & 0.66 &  \textbf{0.71}  & -  \\
\rowcolor[gray]{.92} & \emph{Jaccard} & 0.33 & 0.55 &  0.58  & 0.63  \\
\rowcolor[gray]{.92} & \emph{mWordVec} & 0.28 & 0.53 &  0.56  & 0.59  \\
\rowcolor[gray]{.92} & \emph{pWordVec} & 0.31 & 0.52 &  0.55  & 0.57 \\
\rowcolor[gray]{.92} & \emph{Embed-Num}  &   \textbf{0.39} & \textbf{0.56} & \textbf{0.64} &  \textbf{0.68}  \\
\bottomrule
 \multirow{-11}{*}{\STAB{\rotatebox[origin=c]{90}{ {\normalsize  String data}  }}} & \\
\vspace*{-2em}
\end{tabular}
\end{small} 
\end{center}
\end{table}

In addition to retrieving the correct matches when they exist, useful match-probability estimates should also indicate when a new dataset is unlikely to match any of the previously observed variables' data.  
We consider 100 query datasets randomly chosen from a subset of the datasets in $\mathcal{T}$ which do not have any annotated match in $\mathcal{R}$, and compute the maximum match-probability $m^*$ for each query $\mathcal{D}_*$ when compared against the repository datasets in $\mathcal{R}$.
Avoiding false matches for such a query requires selecting match-likelihood threshold $\ge m^*$, and we evaluate the \emph{calibrated-recall} of each method as the fraction of annotated matches between $\mathcal{T}$ and $\mathcal{R}$ datasets whose estimated match probability exceeds $m^*$.
The final column of Table \ref{tab:retrievenumeric} lists these calibrated-recall values, demonstrating that the predicted  match-probabilities are better calibrated under our approach than the alternative methods.


Among the natural language data, the repository dataset that receives highest probability adjustment value, i.e.\ $\argmax_j g(\mathcal{D}_j)$, stems from a column where 58\% of the values are \texttt{true} and the rest all have the value \texttt{false}.  The OpenML language data contain many other columns also solely comprised of roughly evenly-distributed \texttt{true}/\texttt{false} values, and many of these also appear in the list of the repository datasets that $g_\psi$ assigns the greatest probability adjustment values.  As these boolean columns are labeled with diverse names indicating different unrelated attributes  
(such as whether an animal species is oviparous in one dataset with variable-name \texttt{eggs}), 
our model learns to output a low match probability between such pairs, which is only achievable by ensuring $g_\psi$ outputs large adjustment values for unmatched datasets whose distribution occurs commonly in $\mathcal{R}$.  In contrast, the alternative methods produce small $D_{ij}$ between all pairs of such \texttt{true}/\texttt{false} datasets and thus mistakenly predict a variable-match, demonstrating yet another downside of pure reliance on measures of distributional similarity.

\begin{figure}[bh!]
\begin{minipage}{\textwidth}
\centering
\textbf{(A)} \hspace*{220pt} \textbf{(B)}  \\[-0.03em]
\raisebox{6.5pt}{\includegraphics[width=0.4\textwidth]{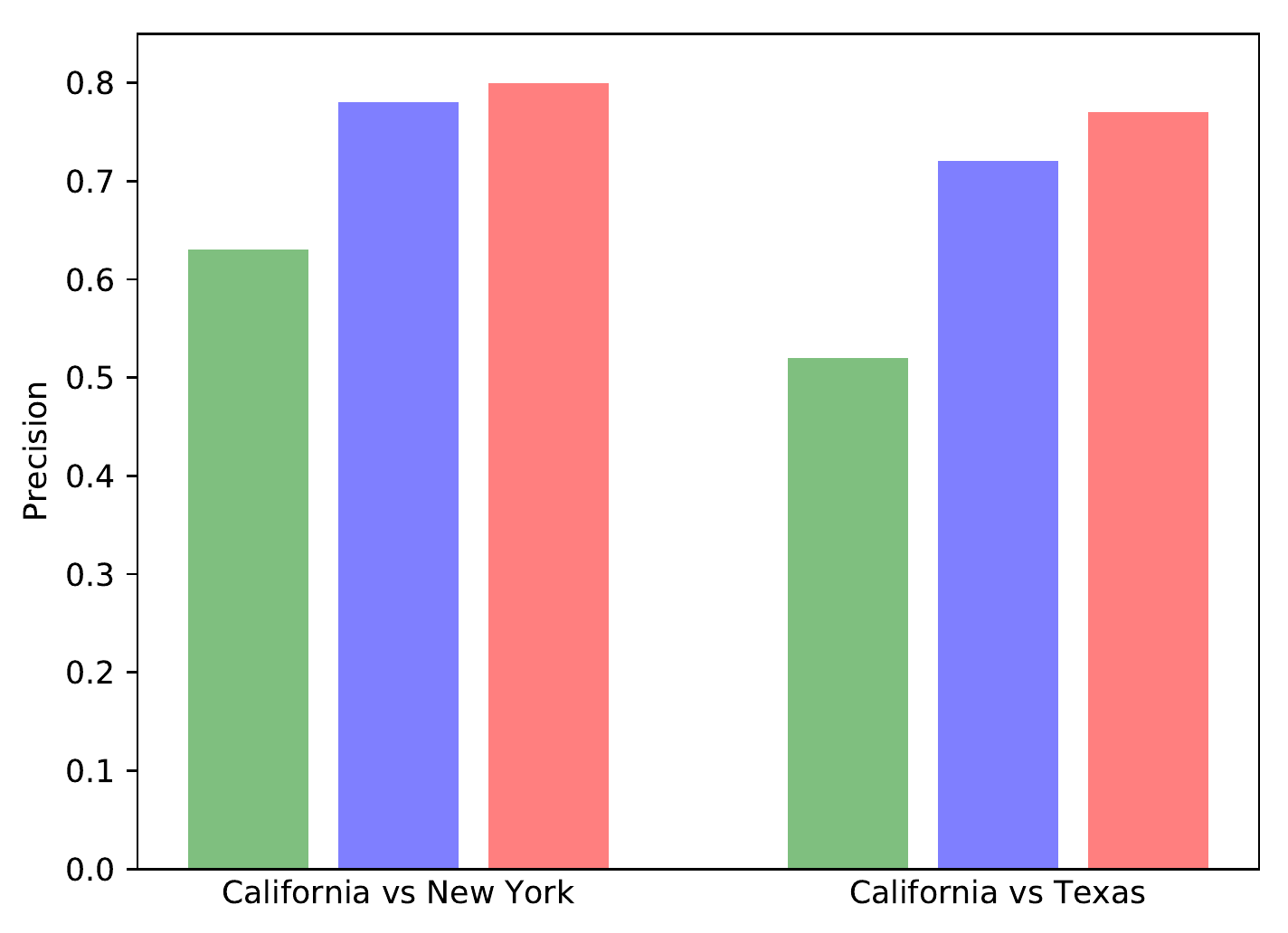}} 
\hspace{0.07\textwidth}
\includegraphics[width=0.405\textwidth]{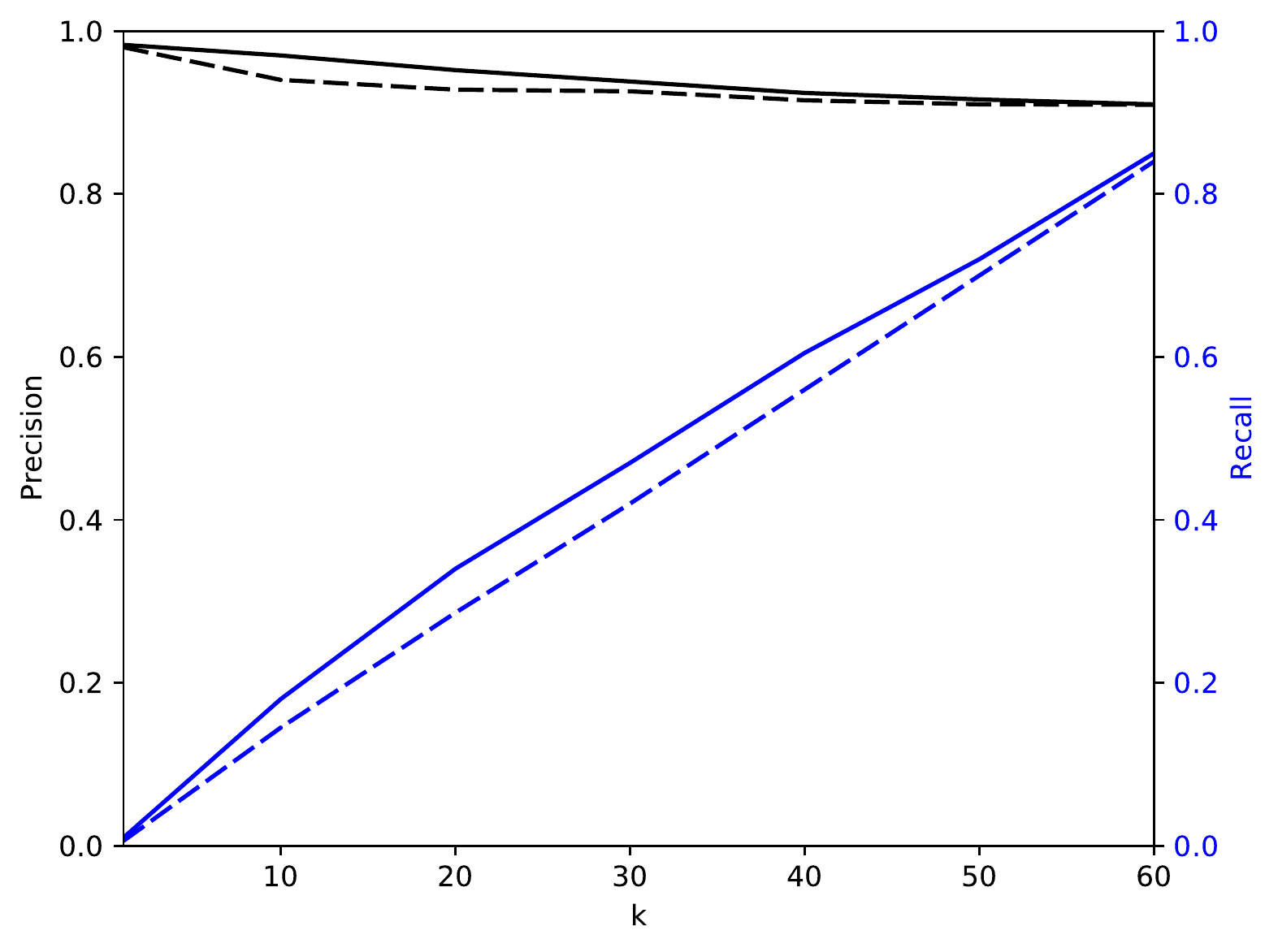}
\vspace*{-0.8em}
\caption{\textbf{(A)} Schema-matching precision between two states' census data for: our OpenML-trained model (red), \emph{FO-LL} (blue), \emph{FO-EP} (green).
\textbf{(B)} Performance in table union search by our OpenML-trained model (solid) and \emph{Ensemble} from \cite{tableunion} (dashed).
}
\label{fig:applications} 
\end{minipage}
\end{figure}

\subsection{Schema Matching}
\label{sec:schemamatching}

 As previous evaluations were based on OpenML column names which might be noisy, we now investigate how well our embedding models trained on the OpenML column names generalize to other datasets where ground truth variable-matches are known.  
First, we apply our OpenML-trained dataset embedding models to the task of instance-based schema matching. We use the PUMS data considered by \cite{Jaiswal13}, in which the task is to match schemas between separate census data tables from New York, California, and Texas.  Evaluation of proposed matchings can be done since the underlying schemas are actually shared across tables, but the methods we apply are not privy to this knowledge (we evaluate on the largest tables studied by \cite{Jaiswal13}: schema-size $=20$). Both numeric and language data types are found within these tables and we use the appropriate corresponding OpenML-pretrained neural network to separately embed each column of each table (with no additional training on the schema matching data).  
Subsequently, we adopt our estimated match probabilities as pairwise column-similarity scores that are fed into the schema-alignment optimization used in \cite{Jaiswal13}, which utilizes local minimization with two-opt switching to find a one-to-one schema alignment. Figure \ref{fig:applications}A shows our pre-trained embeddings are more effective in these schema-matching applications than the two best-performing column-similarity-scoring methods proposed by \cite{Jaiswal13}: \emph{FO-LL} and \emph{FO-EP} (which both rely on distributional similarity between matched variables).

\subsection{Table Union Search}
\label{sec:unionsearch}

Finally, we consider the table union search benchmark of \cite{tableunion}, which involves searching a repository of data tables from the Canadian/UK governments' Open Data portal for tables that can be \emph{unioned} with a given query table (i.e.\ joined row-wise by merging their observations).  Unionability is defined as the condition that a sufficiently large subset of their columns contain measurements of the same underlying variables.
We use the appropriate OpenML-pretrained neural network (for each data type) to separately embed each column of each table (with no additional training on the table union datasets).
The top-performing method of \cite{tableunion} relies on a sophisticated \emph{Ensemble} of multiple manually-crafted statistical similarity measures between columns.
 Our method produces an analogous column similarity-measure (based on our embeddings), from which overall table unionability is evaluated by aggregating similarity across columns via the same max-$c$-alignment procedure \cite{tableunion} use to aggregate their \emph{Ensemble} scores.
Figure \ref{fig:applications}B shows our pre-trained embeddings are quite effective for table union search, producing superior precision/recall compared with the top-performing method in \cite{tableunion}.
While the schema matching and table union evaluations involved vastly different datasets, our OpenML-trained models remain able to perform well, highlighting the broad applicability of these embeddings, as well as the validity of the OpenML column names and our previous OpenML evaluations.

 \renewcommand{\bottomfraction}{.75}
  \renewcommand{\topfraction}{.75}
  \def\bottomnumber{1}

\clearpage 
 \FloatBarrier

\balance{}
\bibliography{IEEEabrv,datasetembeddings.bib} 

\begin{thebibliography}{10}
\providecommand{\url}[1]{#1}
\csname url@samestyle\endcsname
\providecommand{\newblock}{\relax}
\providecommand{\bibinfo}[2]{#2}
\providecommand{\BIBentrySTDinterwordspacing}{\spaceskip=0pt\relax}
\providecommand{\BIBentryALTinterwordstretchfactor}{4}
\providecommand{\BIBentryALTinterwordspacing}{\spaceskip=\fontdimen2\font plus
\BIBentryALTinterwordstretchfactor\fontdimen3\font minus
  \fontdimen4\font\relax}
\providecommand{\BIBforeignlanguage}[2]{{%
\expandafter\ifx\csname l@#1\endcsname\relax
\typeout{** WARNING: IEEEtran.bst: No hyphenation pattern has been}%
\typeout{** loaded for the language `#1'. Using the pattern for}%
\typeout{** the default language instead.}%
\else
\language=\csname l@#1\endcsname
\fi
#2}}
\providecommand{\BIBdecl}{\relax}
\BIBdecl

\bibitem{autostatistician}
C.~Steinrucken, E.~Smith, D.~Janz, J.~Lloyd, and Z.~Ghahramani, ``The automatic
  statistician,'' in \emph{Automatic Machine Learning: Methods, Systems,
  Challenges}, F.~Hutter, L.~Kotthoff, and J.~Vanschoren, Eds.\hskip 1em plus
  0.5em minus 0.4em\relax Springer, 2018, ch.~9, pp. 175--188.

\bibitem{metalearn}
J.~Vanschoren, ``Meta-learning,'' in \emph{Automatic Machine Learning: Methods,
  Systems, Challenges}, F.~Hutter, L.~Kotthoff, and J.~Vanschoren, Eds.\hskip
  1em plus 0.5em minus 0.4em\relax Springer, 2018, ch.~2, pp. 39--68.

\bibitem{openml}
J.~Vanschoren, J.~N. van Rijn, B.~Bischl, and L.~Torgo, ``{OpenML}: Networked
  science in machine learning,'' in \emph{SIGKDD Explorations}, 2013, pp.
  49--60.

\bibitem{Kang03}
J.~Kang and J.~F. Naughton, ``On schema matching with opaque column names and
  data values,'' in \emph{ACM SIGMOD International Conference on Management of
  Data}, 2003, pp. 205--216.

\bibitem{Chen18}
Z.~Chen, H.~Jia, J.~Heflin, and B.~D. Davison, ``Generating schema labels
  through dataset content analysis,'' in \emph{Companion Proceedings of The Web
  Conference ({WWW} ’18)}, 2018, pp. 1515--1522.

\bibitem{Yi18}
Y.~Yi, Z.~Chen, J.~Heflin, and B.~D. Davison, ``Recognizing quantity names for
  tabular data,'' in \emph{International Workshop on Data Search ({ACM} {SIGIR}
  `18)}, 2018, pp. 68--73.

\bibitem{Jaiswal13}
A.~Jaiswal, D.~J.~Miller, and P.~Mitra, ``Schema matching and embedded value
  mapping for databases with opaque column names and mixed continuous and
  discrete-valued data fields,'' \emph{ACM Transactions on Database Systems},
  vol.~38, 2013.

\bibitem{Doan01}
A.~Doan, P.~Domingos, and A.~Y. Halevy, ``Reconciling schemas of disparate data
  sources: A machine-learning approach,'' in \emph{ACM SIGMOD International
  Conference on Management of Data}, vol.~30, no.~2, 2001, pp. 509--520.

\bibitem{dsl}
M.~Pham, S.~Alse, C.~A. Knoblock, and P.~Szekely, ``Semantic labeling: a
  domain-independent approach,'' in \emph{International Semantic Web
  Conference}, 2016, pp. 446--462.

\bibitem{tableunion}
F.~Nargesian, E.~Zhu, K.~Q. Pu, and R.~J. Miller, ``Table union search on open
  data,'' \emph{Proceedings of the {VLDB} Endowment}, vol.~11, no.~7, pp.
  813--825, 2018.

\bibitem{fastneighbors}
M.~Muja and D.~G. Lowe, ``Fast approximate nearest neighbors with automatic
  algorithm configuration.'' \emph{VISAPP}, vol.~2, pp. 331--340, 2009.

\bibitem{neuralstat}
H.~Edwards and A.~Storkey, ``Towards a neural statistician,'' in
  \emph{International Conference on Learning Representations}, 2017.

\bibitem{multivae}
D.~Bouchacourt, R.~Tomioka, and S.~Nowozin, ``Multi-level variational
  autoencoder: Learning disentangled representations from grouped
  observations,'' in \emph{AAAI Conference on Artificial Intelligence}, 2018.

\bibitem{supervisedsemanticlabel}
N.~Ruemmele, Y.~Tyshetskiy, and A.~Collins, ``Evaluating approaches for
  supervised semantic labeling,'' in \emph{Linked Data on the Web}, 2018.

\bibitem{Valera17}
I.~Valera and Z.~Ghahramani, ``Automatic discovery of the statistical types of
  variables in a dataset,'' in \emph{International Conference on Machine
  Learning}, 2017, pp. 3521--3529.

\bibitem{schemablog}
\BIBentryALTinterwordspacing
F.~Greg and P.~Xu, ``Modern approaches to schema matching,'' \emph{DataMade
  Blog}, 2017. [Online]. Available:
  \url{http://datamade.us/blog/schema-matching/}
\BIBentrySTDinterwordspacing

\bibitem{semtyper}
S.~K. Ramnandan, A.~Mittal, C.~A. Knoblock, and P.~Szekely, ``Assigning
  semantic labels to data sources,'' in \emph{European Semantic Web
  Conference}.\hskip 1em plus 0.5em minus 0.4em\relax Springer, 2015, pp.
  403--417.

\bibitem{bayessets}
Z.~Ghahramani and K.~A. Heller, ``Bayesian sets,'' in \emph{Advances in Neural
  Information Processing Systems}, 2006, pp. 435--442.

\bibitem{schemamatchingsurvey}
E.~Rahm and P.~A. Bernstein, ``A survey of approaches to automatic schema
  matching,'' \emph{The VLDB Journal}, vol.~10, no.~4, pp. 334--350, 2001.

\bibitem{mmd}
A.~Gretton, K.~M. Borgwardt, M.~J. Rasch, B.~Sch{\"o}lkopf, and A.~Smola, ``A
  kernel two-sample test,'' \emph{Journal of Machine Learning Research},
  vol.~13, no. Mar, pp. 723--773, 2012.

\bibitem{deepsets}
M.~Zaheer, S.~Kottur, S.~Ravanbakhsh, B.~Poczos, R.~Salakhutdinov, and
  A.~Smola, ``Deep sets,'' in \emph{Advances in Neural Information Processing
  Systems}, 2017.

\bibitem{Chwialkowski15}
K.~P. Chwialkowski, A.~Ramdas, D.~Sejdinovic, and A.~Gretton, ``Fast two-sample
  testing with analytic representations of probability measures,'' in
  \emph{Advances in Neural Information Processing Systems}, 2015, pp.
  1981--1989.

\bibitem{Chopra05}
S.~Chopra, R.~Hadsell, and Y.~LeCun, ``Learning a similarity metric
  discriminatively, with application to face verification,'' in \emph{Computer
  Vision and Pattern Recognition}, vol.~1.\hskip 1em plus 0.5em minus
  0.4em\relax IEEE, 2005, pp. 539--546.

\bibitem{adam}
D.~P. Kingma and J.~Ba, ``Adam: A method for stochastic optimization,'' in
  \emph{International Conference for Learning Representations}, 2015.

\bibitem{facenet}
F.~Schroff, D.~Kalenichenko, and J.~Philbin, ``Facenet: A unified embedding for
  face recognition and clustering,'' in \emph{Computer Vision and Pattern
  Recognition}.\hskip 1em plus 0.5em minus 0.4em\relax IEEE, 2015, pp.
  815--823.

\bibitem{probadjust}
P.~Latinne, M.~Saerens, and C.~Decaestecker, ``Adjusting the outputs of a
  classifier to new a priori probabilities may significantly improve
  classification accuracy: Evidence from a multi-class problem in remote
  sensing,'' in \emph{International Conference on Machine Learning}, 2001.

\bibitem{Batista2004}
G.~E. Batista, R.~C. Prati, and M.~C. Monard, ``A study of the behavior of
  several methods for balancing machine learning training data,'' \emph{ACM
  SIGKDD Explorations Newsletter}, vol.~6, no.~1, pp. 20--29, 2004.

\bibitem{samplingmatters}
C.-Y. Wu, R.~Manmatha, A.~J. Smola, and P.~Kr{\"a}henb{\"u}hl, ``Sampling
  matters in deep embedding learning,'' in \emph{International Conference on
  Computer Vision}.\hskip 1em plus 0.5em minus 0.4em\relax IEEE, 2017.

\bibitem{Horiguchi2017}
S.~Horiguchi, D.~Ikami, and K.~Aizawa, ``Significance of softmax-based features
  in comparison to distance metric learning-based features,''
  \emph{arXiv:1712.10151}, 2017.

\bibitem{pu}
C.~Elkan and K.~Noto, ``Learning classifiers from only positive and unlabeled
  data,'' in \emph{Knowledge Discovery and Data Mining}.\hskip 1em plus 0.5em
  minus 0.4em\relax ACM, 2008.

\bibitem{nalu}
A.~Trask, F.~Hill, S.~E. Reed, J.~Rae, C.~Dyer, and P.~Blunsom, ``Neural
  arithmetic logic units,'' in \emph{Advances in Neural Information Processing
  Systems}, 2018, pp. 8046--8055.

\bibitem{fasttext}
P.~Bojanowski, E.~Grave, A.~Joulin, and T.~Mikolov, ``Enriching word vectors
  with subword information,'' \emph{Transactions of the Association for
  Computational Linguistics}, vol.~5, pp. 135--146, 2017.

\bibitem{lstm}
S.~Hochreiter and J.~Schmidhuber, ``Long short-term memory,'' \emph{Neural
  Computation}, vol.~9, no.~8, pp. 1735--1780, 1997.

\bibitem{jaromatch}
W.~Cohen, P.~Ravikumar, and S.~Fienberg, ``A comparison of string metrics for
  matching names and records,'' in \emph{KDD workshop on data cleaning and
  object consolidation}, 2003.

\bibitem{ks}
W.~J. Conover, \emph{Practical Nonparametric Statistics}.\hskip 1em plus 0.5em
  minus 0.4em\relax New York: John Wiley \& Sons, 1971.

\bibitem{Rahm11}
E.~Rahm, ``Towards large-scale schema and ontology matching,'' in \emph{Schema
  matching and mapping}.\hskip 1em plus 0.5em minus 0.4em\relax Springer, 2011,
  pp. 3--27.

\bibitem{automatch}
J.~Berlin and A.~Motro, ``Database schema matching using machine learning with
  feature selection,'' in \emph{Advanced Information Systems Engineering},
  2002, pp. 452--466.

\end{thebibliography}
\bibliographystyle{IEEEtran}

\end{document}